\documentclass[conference]{IEEEtran}
\IEEEoverridecommandlockouts

\usepackage{verbatim} 
\usepackage{amsfonts} 
\usepackage{algorithm}  
\usepackage{algorithmic}  

\usepackage[table]{xcolor}
\usepackage{graphicx} 
\usepackage{pdfpages} 
\usepackage{epstopdf} 
\usepackage{booktabs} 

\usepackage{cite}
\usepackage{amsmath,amssymb,amsfonts}
\usepackage{algorithmic}
\usepackage{graphicx}
\usepackage{textcomp}
\usepackage{xcolor}
\def\BibTeX{{\rm B\kern-.05em{\sc i\kern-.025em b}\kern-.08em
    T\kern-.1667em\lower.7ex\hbox{E}\kern-.125emX}}
\begin{document}

\title{Adaptive Margin Contrastive Learning for Ambiguity-aware 3D Semantic Segmentation\\
\thanks{$^{*}$ Corresponding author (Email: duanyueqi@tsinghua.edu.cn).
}
}

\author{ 
   \IEEEauthorblockN{Yang Chen$^{1}$, Yueqi Duan$^{2*}$, Runzhong Zhang$^{1}$, and Yap-Peng Tan$^{1}$}
   \\
   \IEEEauthorblockA{$^{1}$ Nanyang Technological University}
   \IEEEauthorblockA{$^{2}$ Tsinghua University}
   }

\maketitle

\begin{abstract}
In this paper, we propose an adaptive margin contrastive learning method for 3D point cloud semantic segmentation, namely AMContrast3D. Most existing methods use equally penalized objectives, which ignore per-point ambiguities and less discriminated features stemming from transition regions. However, as highly ambiguous points may be indistinguishable even for humans, their manually annotated labels are less reliable, and hard constraints over these points would lead to sub-optimal models. To address this, we design adaptive objectives for individual points based on their ambiguity levels, aiming to ensure the correctness of low-ambiguity points while allowing mistakes for high-ambiguity points. Specifically, we first estimate ambiguities based on position embeddings. Then, we develop a margin generator to shift decision boundaries for contrastive feature embeddings, so margins are narrowed due to increasing ambiguities with even negative margins for extremely high-ambiguity points. Experimental results on large-scale datasets, S3DIS and ScanNet, demonstrate that our method outperforms state-of-the-art methods.
\end{abstract}


\section{Introduction}
\label{sec:intro}

3D point cloud semantic segmentation is a task to segment points into semantic coherent regions. The conventional deep learning approaches employ cross-entropy objective to guide model training~\cite{qi2017pointnet},~\cite{qi2017pointnet++},~\cite{hu2020jsenet},~\cite{qian2022pointnext},~\cite{zheng2023learning}. Recently, more scholarly efforts have extended this paradigm by incorporating point-level contrastive objective, which serves as a complementary feature learning strategy to promote compactness within the same semantic regions and dispersion among different semantic regions~\cite{xie2020pointcontrast},~\cite{tang2022contrastive},~\cite{li2022hybridcr},~\cite{jiang2021guided}.

\begin{figure}[t]
\begin{minipage}[b]{1.0\linewidth}
  \centering
  \includegraphics[width=8.8cm]{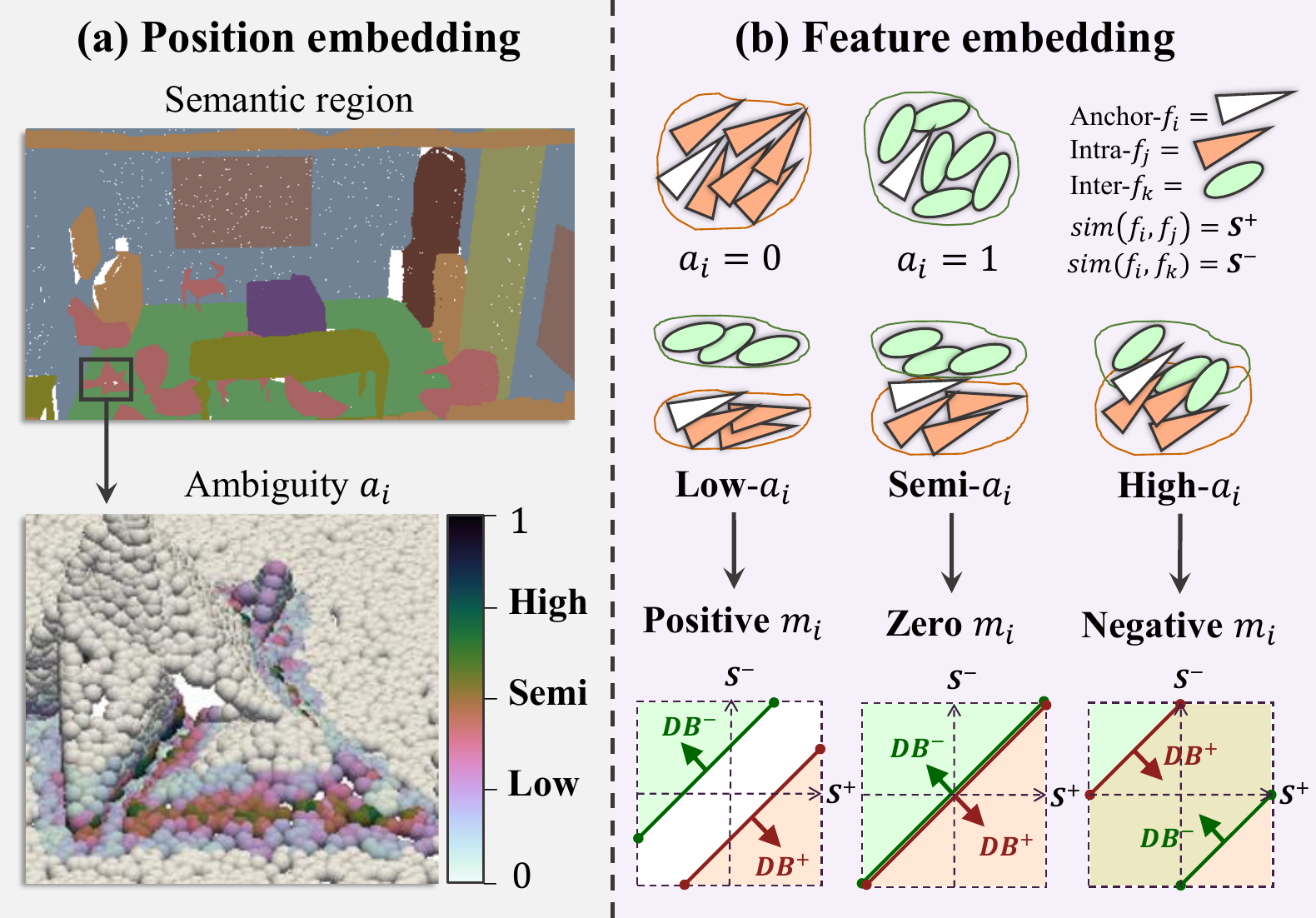}
\end{minipage}
\caption{Adaptive margin from ambiguity. An illustration among (a) position embedding indicates per-point ambiguity $a_i$ colored by a map ranging from $0$ to $1$, and (b) feature embedding yields similarities of intra-pair $S^+$ and inter-pair $S^-$, using ambiguity-aware margin $m_i$ to adjust decision boundaries $DB^+$ and $DB^-$ in contrastive learning, which generates adaptive objectives to benefit embedding learning.}
\label{fig:teaser}
\end{figure}

Despite the effectiveness in enhancing feature discrimination, most prevailing contrastive objectives develop a uniform training difficulty for different points. However, points in transition regions, which commonly interconnect several semantic classes, often exhibit higher sparsity and irregularity compared to those near the object centroid. This inherent disparity introduces inevitable per-point ambiguities that prove challenging for both models and human annotators to distinguish. Consequently, when applying a uniform training difficulty to points in transition regions, the model unavoidably over-prioritizes the segmentation of these points and the optimization of their less discriminated features. This, in turn, results in a lack of attention towards the remaining crucial points, further leading to instability during model training.

Motivated by the disparity of per-point ambiguities, we introduce AMContrast3D, which dynamically tailors training difficulty based on adaptive objectives. Our insight is to assign adaptive objectives to different points according to position embeddings in Fig.~\ref{fig:teaser}. Aligning with 2D tasks that leverage decision margins to heighten training difficulty~\cite{deng2019arcface},~\cite{wang2018cosface},~\cite{li2022towards},~\cite{li2020boosting}, our approach similarly preserves large margins for points with low ambiguities, yet narrows the margins to be smaller even down to negative values for highly ambiguous points in 3D point clouds. In this way, margins positively correlate with training difficulties, and Fig.~\ref{fig:teaser} demonstrates that low-ambiguity, semi-ambiguity, and high-ambiguity points correspond to positive, zero, and negative margins between decision boundaries. By adaptively adjusting the margins for points distributed in different regions, AMContrast3D prevents overfitting to transitional points, concurrently strengthening the feature discrimination capacity and improving overall training robustness.

To this end, we propose an ambiguity estimation framework and an adaptive margin contrastive learning approach integrated into the encoder-decoder network architecture. Specifically, the ambiguity estimation framework first computes per-point ambiguities from position embeddings, where points have compact neighboring points with different semantic labels reflecting high ambiguities. Following ambiguity estimation, the adaptive margin contrastive learning approach constructs an ambiguity-aware margin generator, which dynamically adjusts point-level decision boundaries to regularize intra-class and inter-class feature embeddings from decoder layers. To our knowledge, we are the first to develop negative margins in a contrastive objective for 3D tasks. Extensive experiments on two large-scale datasets demonstrate that AMContrast3D enhances feature discrimination and outperforms baseline models. The ablation study further validates the effectiveness of our proposed method.

\section{Related Work}
\label{sec:rw}

\subsection{Point-based Semantic Segmentation}

PointNet~\cite{qi2017pointnet} pioneers the 3D semantic segmentation, which directly works on irregular point clouds. This network processes individual points with shared MLPs to aggregate global features. However, its performances are limited because of the lack of considering local spatial relations in the point cloud structure. Following PointNet, PointNet++~\cite{qi2017pointnet++} develops a hierarchical spatial structure on local regions with MLPs, termed the set abstraction block. In MLPs-based philosophy, follow-up methods develop novel modules~\cite{zhang2024geoauxnet,liu2020closer}. PointNeXt~\cite{qian2022pointnext} revisits training and scaling strategies, tweaking the set abstraction block. The recently proposed method, PointMetaBase~\cite{lin2023meta}, designs building blocks into four meta functions for point cloud analysis. Compared with convolutional kernels~\cite{thomas2019kpconv},~\cite{xu2021paconv},~\cite{liu2020semantic}, graph structures~\cite{landrieu2018large},~\cite{qian2021pu},~\cite{tao2022seggroup}, and transformer architectures~\cite{zhao2021point},~\cite{park2022fast}, the highly-optimized MLPs are conceptually simpler to reduce computational and memory costs and achieve results on par or better.

\subsection{Contrastive Learning}

Contrastive learning is widely used to pull together feature embeddings from the same class and push away the feature embeddings from different classes~\cite{gutmann2010noise},~\cite{oord2018representation},~\cite{khosla2020supervised},~\cite{ong2023quad}. Works that follow this path design various contrastive objectives on 3D tasks in unsupervised approach~\cite{xie2020pointcontrast}, weakly-supervised approach~\cite{li2022hybridcr}, semi-supervised approach~\cite{jiang2021guided} and supervised approach~\cite{tang2022contrastive}. However, they only conduct fixed contrast on feature embeddings while ignoring adaptive ambiguities from position embeddings.

\subsection{Margin-based Training Objective}

The typical networks use the cross-entropy objective during training. 2D tasks have witnessed a surge regarding decision margins to adjust the objective and strengthen the discriminating power~\cite{deng2019arcface},~\cite{wang2018cosface}. Recent works propose dynamic margins that are proven effective~\cite{li2022towards},~\cite{li2020boosting}, yet they are mostly constrained on positive margins to heighten objectives. This direction is essentially an under-explored aspect of 3D tasks. Meanwhile, considering the intrinsic properties of point clouds, one-sided margins are restrictive. Our method deviates from one-sided margins by exploring adaptive margins involving a diversity of positive, zero, and negative values to generate adaptive objectives.

\section{Method}
\label{sec:method}

\begin{figure*}[t]
\begin{minipage}[b]{1.0\linewidth}
  \centering
  \includegraphics[width=18cm]{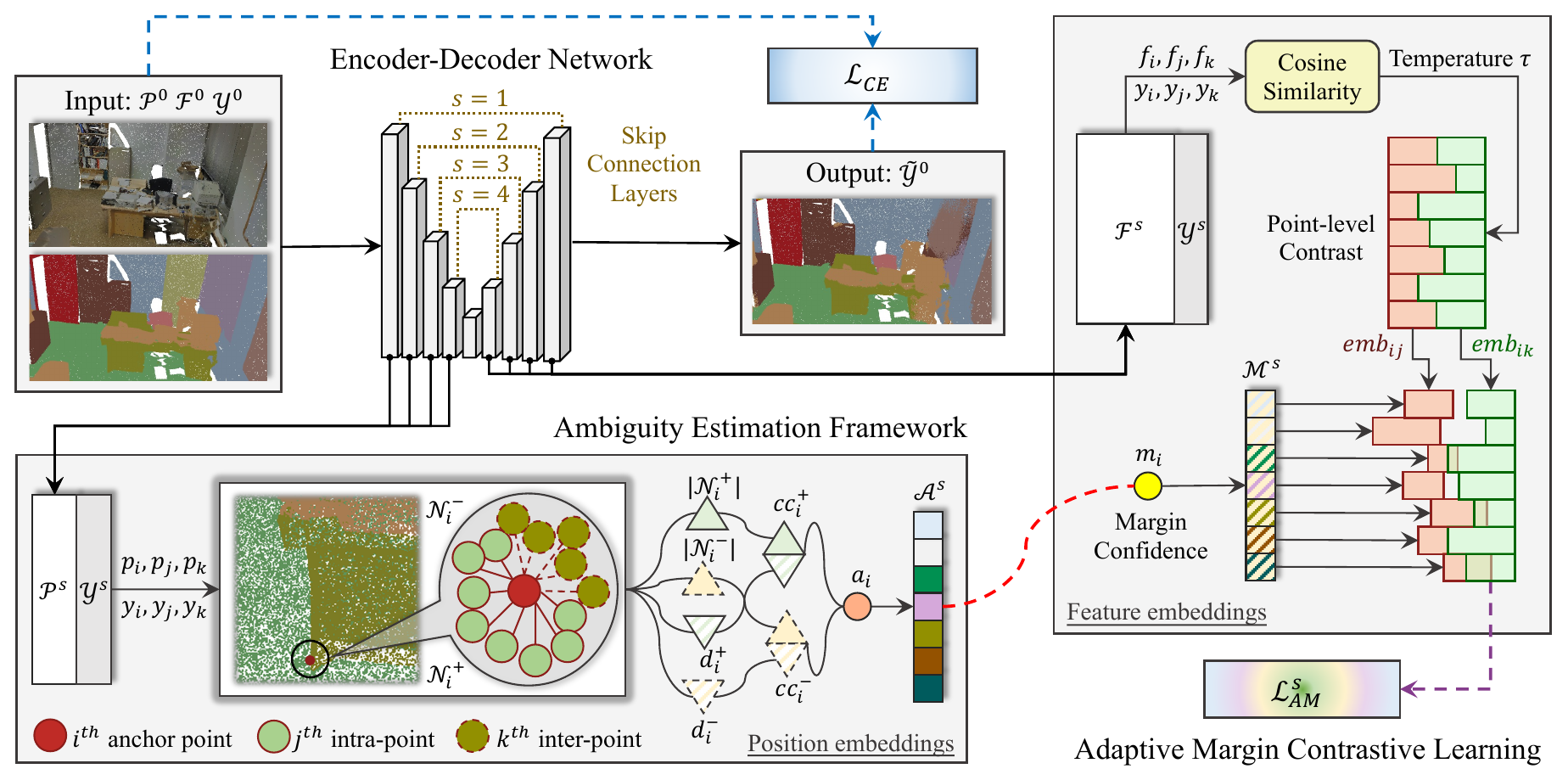}
\end{minipage}
\caption{The AMContrast3D with encoder-decoder network architecture. In the ambiguity estimation framework following the $s^{th}$ encoder layer, we infer the ambiguity $a_i \in \mathcal{A}^s$ for each $i^{th}$ point by encoding position embeddings $p_i,p_j, p_k \in \mathcal{P}^s$ based on the $j^{th}$ intra-points in neighborhood $\mathcal{N}_i^+$ and the $k^{th}$ inter-points in neighborhood $\mathcal{N}_i^-$. We reformulate $a_i$ into adaptive ambiguity-aware margins $m_i \in \mathcal{M}^s$. These margins target feature embeddings $f_i,f_j,f_k \in \mathcal{F}^s$ for each corresponding decoder layer to dynamically adjust decision boundaries during contrastive learning. Through the adaptive margin contrastive learning, our method automatically regulates training difficulties across different parts of the point clouds, particularly ensuring more stabilized training for high-ambiguity points in transition regions containing different semantic classes.}
\label{fig:framework}
\end{figure*}

\subsection{Problem Formulation}
\label{sec:method_FO}
3D semantic segmentation aims to categorize points to the specific classes within a point cloud scene. A point cloud input is a set of 3D points with $\{(p_i, f_i) | i=1,...,n\}$, where $p_i \in \mathbb{R}^3$ is the position and $f_i \in \mathbb{R}^D$ is the $D$-dim feature of the $i^{th}$ point. During inference, the output is the predicted label $\tilde{y}_i \in \mathbb{R}^C$ for each of the $n$ points based on $C$ semantic classes in a dataset, and the ground truth label is $y_i \in \mathbb{R}^C$.

As in Fig.~\ref{fig:framework}, we introduce an adaptive margin contrastive learning method, referred to as AMContrast3D, tailored for embedding learning of ambiguous points. The following subsections comprehensively explain the key components, including the ambiguity estimation framework, ambiguity-aware margin generator, and contrastive optimization.

\subsection{Ambiguity Estimation Framework}
\label{sec:method_AEF}

The ambiguity estimation framework generates per-point ambiguities by exploring the positional relations within local regions, aiming to indicate whether a point is ambiguously challenging to segment and determine its training difficulty.

Given an $i^{th}$ point, we compute Euclidean distances to define its $K$-nearest neighbor points in a set $\mathcal{N}_i^+$. Within $\mathcal{N}_i^+$, most $j^{th}$ neighbor points are intra-points with the same semantic label as $y_j = y_i$, resulting in unambiguous embedding learning. If some $k^{th}$ neighbor points are inter-points with $y_k \neq y_i$, we reallocate them to a new set as $\mathcal{N}_i^-$, which means the $i^{th}$ point is in a transition region, encountering both a positive impact from the intra-class and a negative impact from the inter-class. Intuitively, under a fixed neighboring  size as $K = |\mathcal{N}_i^+| + |\mathcal{N}_i^-|$, larger $|\mathcal{N}_i^-|$ negatively reflects higher ambiguity. Inspired by the closeness centrality in graphs~\cite{veremyev2019finding}, which measures the average inverse distance of a node to all other nodes, we further reconsider a point as a node in an unconnected graph and design two kinds of closeness centrality by position embeddings $p_i, p_j, p_k$ as:
\begin{eqnarray}
\label{eq:cc}
cc_i^+ = (\frac{\sum_{j=1}^{|\mathcal{N}_i^+|}(p_i - p_j)^2}{|\mathcal{N}_i^+|})^{-1} = \frac{|\mathcal{N}_i^+|}{d_i^+}, \\
cc_i^- = (\frac{\sum_{k=1}^{|\mathcal{N}_i^-|}(p_i - p_k)^2}{|\mathcal{N}_i^-|})^{-1} = \frac{|\mathcal{N}_i^-|}{d_i^-}, 
\end{eqnarray}
where the $i^{th}$ point has different compactness with all intra-points as $cc_i^+$ and with all inter-points as $cc_i^-$. Closeness centrality highly correlates with point importance to reflect its compact relation within a neighborhood. Significantly discrepant $cc_i^+$ and $cc_i^-$ are derived from various point numbers and irregular position embeddings of intra-points and inter-points. We find that such a discrepancy of a point can be formulated as a paired subtraction between $cc_i^+$ and $cc_i^-$ in a local neighborhood, which indicates a rational proxy for the ambiguous level. Therefore, we leverage $cc_i^+ - cc_i^-$ into a monotonic decreasing curve, which is formulated as an inverse sigmoid function $\mathcal{G}(cc_i^+, cc_i^-) \in (0,1)$:
\begin{eqnarray}
\label{eq:G}
\mathcal{G}(cc_i^+, cc_i^-) = \frac{1}{1+exp(\beta(cc_i^+-cc_i^-))},
\end{eqnarray}
where $\beta$ is a tuning parameter. Concretely, a large $cc_i^+$ and a small $cc_i^-$ present on a point with low ambiguity approaches $0$; on the contrary, high ambiguity approaches $1$. The minimum $|\mathcal{N}_i^+|$ is $1$, which is a possible circumstance meaning that a neighborhood only contains $1$ intra-point as the $i^{th}$ point itself, and the other points are inter-points. We define such a point as extremely ambiguous, with the highest value equal to $1$. To consider all circumstances, a piece-wise function estimates ambiguity $a_i \in \mathcal{A}^s$ in each layer as:
\begin{eqnarray}
\label{eq:A}
a_i = 
\begin{cases}
0 & \text{if } |\mathcal{N}_i^+| = K, \\
\mathcal{G}(cc_i^+, cc_i^-) & \text{if } 1 < |\mathcal{N}_i^+| < K, \\
1 & \text{if } |\mathcal{N}_i^+| = 1. 
\end{cases}
\end{eqnarray}

\begin{figure}[t]
\begin{minipage}[b]{1.0\linewidth}
  \centering  
  \includegraphics[width=8.8cm]{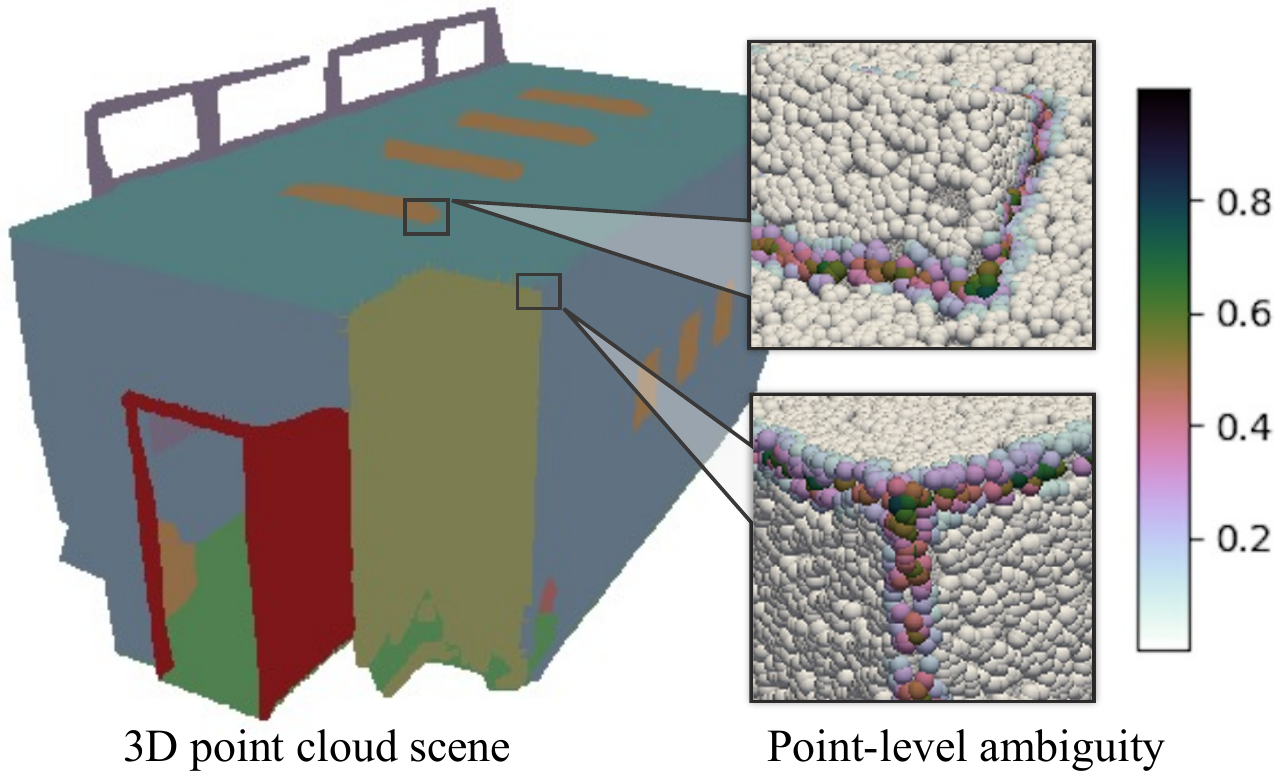}
\end{minipage}
\caption{Ambiguity visualization. A 3D point cloud scene is categorized by different semantic classes. We visualize the point-level ambiguity for each point, where the color from white to black indicates various ambiguity levels ranging in $[0,1]$.}
\label{fig:aieq}
\end{figure}

Fig.~\ref{fig:aieq} visualizes the ambiguity $a_i$ in a point cloud scene. Low-ambiguity, semi-ambiguity, and high-ambiguity points with $a_i \in (0,1]$ are surrounded by various numbers of inter-points in transition regions. We focus on these points to assign point-level contrast dynamically to stabilize the training.

\subsection{Adaptive Margin Contrastive Learning}
\label{sec:method_CL}

Contrastive learning encourages intra-class compactness and inter-class separability. We extend its supervised approach with ambiguity awareness to benefit embedding learning.

\textbf{Revisiting Contrastive Learning.} Following the setup of a temperature parameter $\tau$ controlling the contrast~\cite{khosla2020supervised}, a supervised contrastive objective for the $i^{th}$ point is
\begin{eqnarray}\label{eq:LOSS_CL}
    - log\frac{\sum_{j=1}^{|\mathcal{N}_i^+|}exp(\frac{sim(f_i, f_j)}{\tau})}{\sum_{j=1}^{|\mathcal{N}_i^+|}exp(\frac{sim(f_i, f_j)}{\tau}) + \sum_{k=1}^{|\mathcal{N}_i^-|}exp(\frac{sim(f_i, f_k)}{\tau})},
\end{eqnarray} 
which intensifies discrimination on feature similarities, maximizing intra-pair $sim(f_i, f_j)$ while minimizing inter-pair $sim(f_i, f_k)$. Eq.~(\ref{eq:LOSS_CL}) shares a common ground with cross-entropy objective~\cite{he2020momentum}, and decision margins can modify the cross-entropy objective~\cite{deng2019arcface},~\cite{wang2018cosface},~\cite{li2022towards}. Thus, decision boundaries $DB^+$ for intra-pairs and $DB^-$ for inter-pairs in Eq.~(\ref{eq:LOSS_CL}) are
\begin{eqnarray}\label{eq:DB_0}
\begin{aligned}
DB^+: sim(f_i, f_j) - sim(f_i, f_k) \ge 0, \\
DB^-: sim(f_i, f_j) - sim(f_i, f_k) \le 0.
\end{aligned}
\end{eqnarray}

Eq.~(\ref{eq:LOSS_CL}) and Eq.~(\ref{eq:DB_0}) pose two limitations: 1) The margin is $0$, which means $DB^+$ and $DB^-$ are adjacent without discrimination. 2) The objective only targets feature embeddings in $\mathcal{F}^s$ but completely disregards position embeddings in $\mathcal{P}^s$.

\textbf{Ambiguity-aware Margin Generator.} We address these limitations by margins. Intuitively, a fixed positive margin directly generates expansion between $DB^+$ and $DB^-$, forcing all points to reach a complicated contrastive objective identically. Since individual points with various ambiguities require adaptive training objectives, for the generator to make use of the ambiguities, we explicitly inject $a_i \in (0,1]$ as margin confidence to generate adaptive margin $m_i$ as:
\begin{eqnarray}
\label{eq:AMG_mi}
m_i = \mu \cdot a_i+ \nu,
\end{eqnarray}
where $\mu$ and $\nu$ are the scale and bias parameters, respectively. As shown in Fig.~\ref{fig:teaser}, the point-level and ambiguity-aware generator dynamically controls feature discrimination between intra-class and inter-class. Margins involve positive, zero, and negative values in a principled manner: positive margins with penalized separations between $DB^+$ and $DB^-$ heighten objectives for low-ambiguity points; zero margins remain for semi-ambiguity points; negative margins allow moderate feature mixtures with easy objectives, properly degrading the training difficulty for high-ambiguity points. Concretely, we measure similarity metrics using cosine similarity, $i.e.,$ the intra-pair similarity is $sim(f_i,f_j)= \frac{f_i \cdot f_j}{\left \| f_i \right \| \left \| f_j \right \|} \in [-1,1]$. As a discriminative hyperplane, margin $m_i$ dynamically shifts decision boundaries as:
\begin{eqnarray}\label{eq:DB_mi}
\begin{aligned}
DB^+: sim(f_i,f_j) - sim(f_i,f_k) \ge m_i, \\
DB^-: sim(f_i,f_j) - sim(f_i,f_k) \le m_i.
\end{aligned}
\end{eqnarray}

Fig.~\ref{fig:framework} illustrates a red dotted line from ambiguities $a_i \in \mathcal{A}^s$ to margins $m_i \in \mathcal{M}^s$, connecting position embeddings in $\mathcal{P}^s$ and feature embeddings in $\mathcal{F}^s$. Then, margins modify the contrastive objective to provide adaptive training difficulty for each point, stabilizing the overall training.

\textbf{Contrastive Optimization.} We optimize the contrastive learning by encouraging the intra-similarity $sim(f_i, f_j)$ to be larger than the inter-similarity $sim(f_i, f_k)$ plus the margin $m_i$. To satisfy Eq.~(\ref{eq:DB_mi}), we generalize the contrastive embeddings $emb_{ij}$ for intra-pairs and $emb_{ik}$ for inter-pairs as:
\begin{eqnarray}
\label{eq:EMB_mi}
\begin{aligned}
    emb_{ij} = exp(\frac{sim(f_i, f_j) - m_i}{\tau}), \\
    emb_{ik} = exp(\frac{sim(f_i, f_k)}{\tau}),
\end{aligned}
\end{eqnarray}
where temperature $\tau$ controls the contrast level. Algorithm~\ref{code:3AM} explains the generating procedure of $emb_{ij}$ and $emb_{ik}$, from which we develop an adaptive margin contrastive objective as $\mathcal{L}_{AM}^s$. Suppose $n^s$ is the total number of low-$a_i$, semi-$a_i$, and high-$a_i$ points in the $s^{th}$ layer, the objective $\mathcal{L}_{AM}^s$ is
\begin{eqnarray}
\label{eq:LOSS_DMCL}
\begin{aligned}
    \mathcal{L}_{AM}^s = \frac{1}{n^s}\sum_{i=1}^{n^s} - log\frac{\sum_{j=1}^{|\mathcal{N}_i^+|}emb_{ij}}{\sum_{j=1}^{|\mathcal{N}_i^+|}emb_{ij} + \sum_{k=1}^{|\mathcal{N}_i^-|}emb_{ik}}.
\end{aligned}
\end{eqnarray}

This objective maximizes $emb_{ij}$ and minimizes $emb_{ik}$, making an anchor point to be similar to intra-points compared to inter-points. The segmentation model is under joint training by $\mathcal{L}_{AM}^s$ and the cross-entropy objective $\mathcal{L}_{CE}$. $\mathcal{L}_{CE}$ regularizes the prediction $\tilde{y}_i$ based on the ground truth $y_i$ for each of the $n$ points within $C$ semantic classes in point clouds. With a balanced parameter $\lambda$, the overall objective $\mathcal{L}$ is
\begin{eqnarray}
\label{eq:LOSS-T}
\mathcal{L} = \lambda \mathcal{L}_{CE} + (1-\lambda) \sum_{s}\mathcal{L}_{AM}^s, \\
\text{with}\quad \mathcal{L}_{CE}=\frac{-1}{n \cdot C}\sum_{i=1}^{n}\sum_{c=1}^{C}y_{i,c} \cdot log\frac{exp(\tilde{y}_{i,c})}{\sum_{c=1}^{C}exp(\tilde{y}_{i,c})}.
\end{eqnarray}

\begin{algorithm}[t]
    \caption{AMContrast3D of the $i^{th}$ point in the $s^{th}$ layer.} \label{code:3AM}
    \begin{algorithmic}[1]%
    \REQUIRE  $p_i \in \mathcal{P}^s$, $f_i \in \mathcal{F}^s$, $y_i \in \mathcal{Y}^s$, size $K$, temperature $\tau$
    \ENSURE $a_i \in \mathcal{A}^s, m_i \in \mathcal{M}^s, emb_{ij}, emb_{ik}$
    \STATE Neighbor points $\leftarrow p_i$ \COMMENT{\texttt{Position space} $\mathcal{P}^s$}
    \FOR{$j,k$ in $K$}
        \STATE Compute $|\mathcal{N}_i^+|$, $d_i^+$ \COMMENT{\texttt{Intra:} $y_i = y_j$}
        \STATE Compute  $|\mathcal{N}_i^-|$, $d_i^-$ \COMMENT{\texttt{Inter:} $y_i \neq y_k$}   
    \ENDFOR
    \IF{$\mathcal{N}_i^+,\mathcal{N}_i^- \neq \emptyset$}
        \STATE Generate $cc_i^+$ and $cc_i^-$
    \ENDIF
    \STATE Update $\mathcal{A}^s$ from estimated ambiguity $a_i$
    \STATE Update $\mathcal{M}^s$ from margin $m_i$ with $a_i$ awareness
    \STATE $(f_i, f_j), (f_i, f_k)  \leftarrow f_i, f_j, f_k$  \COMMENT{\texttt{Feature space} $\mathcal{F}^s$}
    \FOR{$j,k$ in $K$}
        \STATE $emb_{ij} \leftarrow sim(f_i, f_j), \tau, m_i$ \COMMENT{\texttt{Intra:} $y_i = y_j$}
        \STATE $emb_{ik} \leftarrow sim(f_i, f_k), \tau$ \COMMENT{\texttt{Inter:} $y_i \neq y_k$} 
    \ENDFOR
    \end{algorithmic}
\end{algorithm}

\section{Experiments}
\label{sec:exp}

\subsection{Implementation Details}

We adopt a deep MLPs-based encoder-decoder backbone, PointNeXt~\cite{qian2022pointnext}, which has $4$ skip connection layers from encoder to decoder and contains a stem MLP with a channel size of $64$, InvResMLP blocks with a number as $(3,6,3,3)$, and SA blocks from PointNet++~\cite{qi2017pointnet++}. Based on the label mining strategy from~\cite{tang2022contrastive}, we integrate label embeddings in each layer. The neighboring size $K$ is $24$. Parameters $\beta$ and $\lambda$ are $0.04$ and $0.1$, respectively, and we use an initial learning rate of $0.01$ with $100$ epochs for a training episode.

\subsection{Performance Comparison}

\begin{table}[t]
\begin{center}
\caption{3D semantic segmentation on S3DIS (Area 5).} \label{tab:s3dis}
\begin{tabular}{l|c|c|c}
  \toprule
  Method & OA ($\%$) & mACC ($\%$) & mIoU ($\%$)
  \\
  \midrule
  PointNet~\cite{qi2017pointnet}  & - & 49.0 & 41.1\\
  PointNet++~\cite{qi2017pointnet++}  & 83.0 & - & 53.5\\
  PCT~\cite{guo2021pct}  & - & 67.7 & 61.3 \\
  SPG~\cite{landrieu2018large}  & 85.5 & 73.0 & 62.1\\
  KPConv~\cite{thomas2019kpconv}  & - & 72.8 & 67.1\\
  PAConv~\cite{xu2021paconv}  & - & 73.0 & 66.6\\
  JSENet~\cite{hu2020jsenet}  & - & - & 67.7\\
  CBL~\cite{tang2022contrastive}  & 90.6 & 75.2 & 69.4\\
  PointTransformer~\cite{zhao2021point}  & 90.8 & 76.5 & 70.4\\
  PointMetaBase~\cite{lin2023meta} & 90.8 & - & 71.3\\
  PointNeXt~\cite{qian2022pointnext}  & 90.6 & - & 70.5\\
  \midrule
  AMContrast3D (ours)  & \textbf{91.1} & \textbf{77.1} & \textbf{71.8}\\
  \bottomrule
\end{tabular}
\end{center}
\end{table}

\begin{table}[t]
\begin{center}
\caption{3D semantic segmentation on ScanNet.} \label{tab:scannet}
\begin{tabular}{l|c|c}
  \toprule
  Method  & mIoU (Val) ($\%$) & mIoU (Test) ($\%$)
  \\
  \midrule
  PointNet++~\cite{qi2017pointnet++} & 53.3 & 33.9 \\
  PointConv~\cite{wu2019pointconv}  & 61.0 & 55.6 \\
  KPConv~\cite{thomas2019kpconv}  & 69.2 & 68.6 \\
  JSENet~\cite{hu2020jsenet}  & -  & 69.9\\
  CBL~\cite{tang2022contrastive}  & -  & 70.5\\
  FastPointTransformer~\cite{park2022fast} & 72.1 & - \\
  PointMetaBase~\cite{lin2023meta} & \textbf{72.8} & 71.4 \\
  PointNeXt~\cite{qian2022pointnext}  & 71.5 & 71.2 \\
  \midrule
  AMContrast3D (ours)  & 72.5 & \textbf{72.6} \\
  \bottomrule
\end{tabular}
\end{center}
\end{table}

\begin{figure}[t]
\begin{minipage}[b]{1.0\linewidth}
  \centering
  \includegraphics[width=8.8cm]{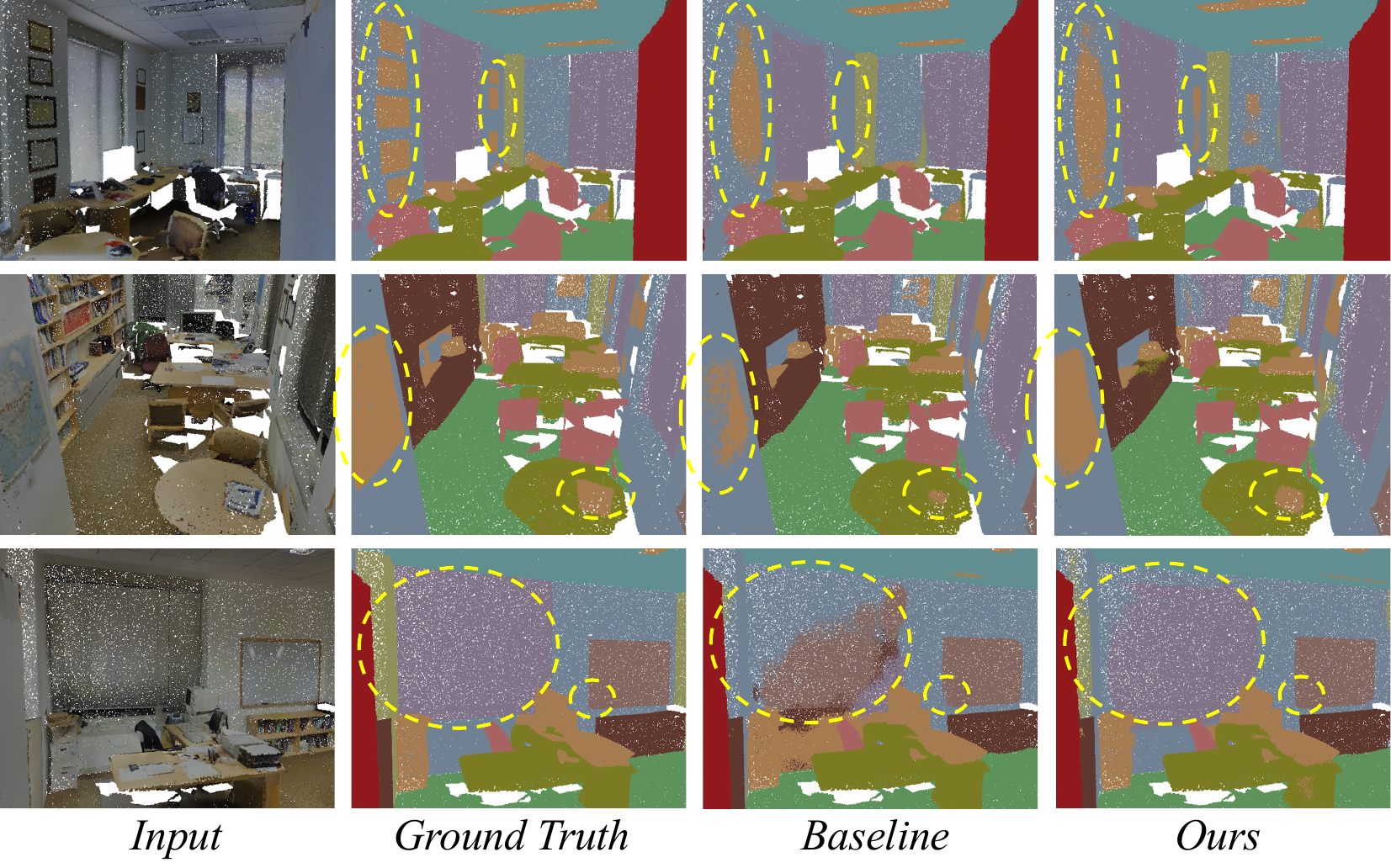}
\end{minipage}
\caption{Visualization results on S3DIS (Area 5). The images from left to right are the input scene, ground truth of semantic labels, results predicted by PointNeXt, and our method.}
\label{fig:s3dis}
\end{figure}

\begin{figure}[t]
\begin{minipage}[b]{1.0\linewidth}
  \centering
  \includegraphics[width=8.8cm]{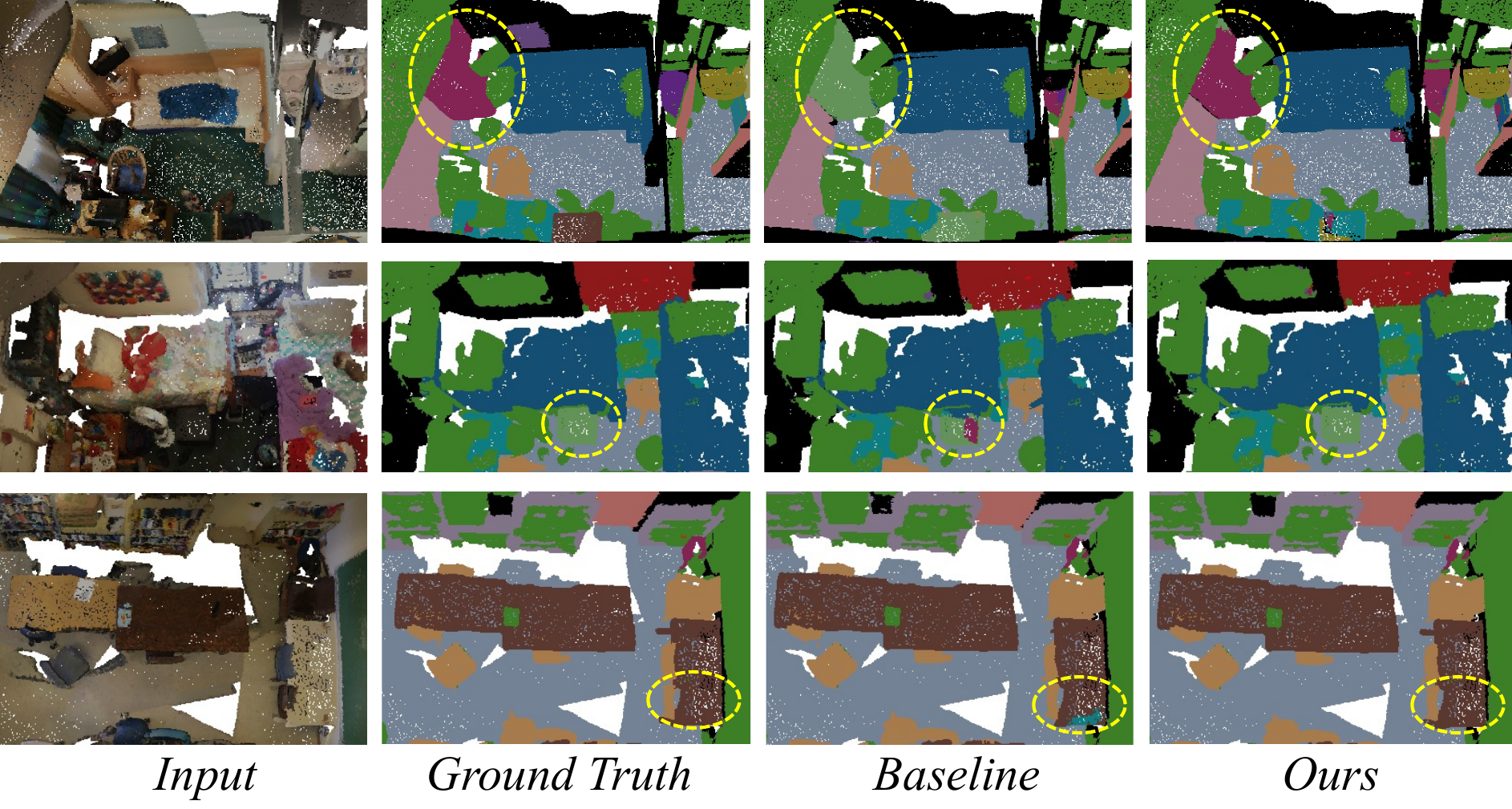}
\end{minipage}
\caption{Visualization results on ScanNet. The images from left to right are the input scene, ground truth of semantic labels, results predicted by PointNeXt, and our method.}
\label{fig:scannet}
\end{figure}

We conduct experiments on two large-scale scene datasets, S3DIS~\cite{armeni2017joint} and ScanNet~\cite{dai2017scannet}. The evaluation metrics contain mean intersection over union (mIoU), overall accuracy (OA), and the mean accuracy within each class (mACC).

\textbf{S3DIS Semantic Segmentation.} S3DIS~\cite{armeni2017joint} covers $271$ rooms in $6$ areas with total semantic classes $C$ as $13$. We take Area 5 for inference and others for training. Input points are downsampled with $n$ as $24000$ per batch. The margin $m_i$ adjusts for each point with scale parameter $\mu$ as $-1$ and bias parameter $\nu$ as $0.5$. Thus, points with low $a_i \in (0,0.5)$ lead to $m_i>0$ for expanded decision boundaries; points with $a_i = 0.5$ have $m_i=0$ for adjacent decision boundaries; points with high $a_i \in (0.5,1]$ lead to $m_i<0$ with mixed decision boundaries. The temperature $\tau$ is $0.3$. Table~\ref{tab:s3dis} provides quantitative results, demonstrating that our method improves the baseline, achieving leading performances of $91.1\%$ in OA, $77.1\%$ in mACC, and $71.8\%$ in mIoU.

\textbf{ScanNet Semantic Segmentation.} ScanNet~\cite{dai2017scannet} is annotated with classes $C$ as $20$. It contains $1613$ cluttered scans, which are split into $1201$ training scans, $312$ validation scans, and $100$ test scans. The point number $n$ is $64000$ per batch. The setting of $\tau$ is $0.5$. Point-level margin $m_i$ has $\mu$ as $-1$ and $\nu$ as $0.6$. Thereby, points with $a_i \in (0,0.6)$, $a_i = 0.6$, and  $a_i \in (0.6,1]$ lead to positive, zero, and negative $m_i$, respectively. Our method achieves mIoU of $72.6\%$, which performs better than many competitors, as shown in Table~\ref{tab:scannet}. 

Our method attains significant improvements against the baseline that performs better than PointNeXt by $1.3\%$ for S3DIS and $1.4\%$ for ScanNet in mIoU. Fig.~\ref{fig:s3dis} and Fig.~\ref{fig:scannet} visually present the results in point clouds of two datasets. We leverage concise and straightforward MLP structure, which is computationally efficient compared with the convolutional structures~\cite{thomas2019kpconv, xu2021paconv} and the transformer structures~\cite{zhao2021point, park2022fast}, and it outperforms these complicated methods during inference. Meanwhile, our method has competitiveness with the recently proposed MLPs-based method~\cite{lin2023meta}.

\subsection{Ablation Study}

To evaluate the effectiveness of the key component, margin $m_i$, we conduct an ablation study on the S3DIS dataset with different settings of $m_i$ in Table~\ref{tab:margin}. The first two rows are constant margins, and the remaining rows are adaptive margins. Constant $m_i$ generates uniform contrastive objectives without considering the disparity of ambiguities, which can not capture sufficient context for high-ambiguity points, and the performance significantly drops. The best mIoU result is achieved with $\mu=-1$ and $\nu=0.5$ that uses adaptive $m_i$ controlled by $a_i \in (0,1]$. In this case, margins cover positive, zero, and negative values to determine the expansion or shrinkage of decision boundaries. This ablation suggests that negative values are essential for ambiguity-aware margins.

\begin{table}[t]
\begin{center}
\caption{Results on the ablation study of adaptive margins involving positive ($0 \uparrow$), zero ($0$), and negative ($0 \downarrow$) values.} \label{tab:margin}
\begin{tabular}{c|c|c|c|c|c|c}
  \toprule
  $\mu$ & $\nu$ & $m_i$ & $0 \uparrow$ & $0$ & $0 \downarrow$ & mIoU ($\%$)
  \\
  \midrule
  $0$ & $0$ & $0$ &  & $\checkmark$ &  & 70.5 \\
  $0$ & $0.5$ & $0.5$ & $\checkmark$ &  &  & 69.7 \\
  \rowcolor{blue!7} $1$ & $0$ & $a_i$ & $\checkmark$ &  &  & 70.1 \\
  \rowcolor{blue!7} $-1$ & $1$ & $1 - a_i$ & $\checkmark$ & $\checkmark$ &  & 70.6 \\
  \rowcolor{green!7} $-1$ & $0.5$ & $0.5 - a_i$ & $\checkmark$ & $\checkmark$ & $\checkmark$ & \textbf{71.8} \\
  \rowcolor{blue!7} $-1$ & $0.5$ & $max(0, 0.5 - a_i)$ & $\checkmark$ & $\checkmark$ &  & 70.5 \\
  \bottomrule
\end{tabular}
\end{center}
\end{table}

\section{Conclusion}
In this paper, we propose an ambiguity estimation framework and an ambiguity-aware margin generator to achieve adaptive contrastive objectives for 3D semantic segmentation. Extensive experiments on two datasets demonstrate that our method boosts the segmentation performance, and the ablation empirically validates our claim that additional negative margins benefit embedding learning during training. Our method inspires new insights to rethink per-point ambiguities in point clouds and explore ambiguity-aware attention during inference in future research.

\section{Acknowledgment}

This work is supported in part by the National Research Foundation of Singapore under the NRF Medium Sized Centre Scheme (CARTIN). Any opinions, findings and conclusions expressed in this material are those of the authors and do not reflect the views of National Research Foundation, Singapore. This work is also supported in part by the National Natural Science Foundation of China under Grant 62206147.

\bibliographystyle{IEEEbib} 

\end{document}